\documentclass[11pt,a4paper]{article}
\usepackage[utf8]{inputenc}  
\usepackage{amsmath}
\usepackage[]{iclr}
\let\oldhat\hat
\renewcommand{\hat}[1]{\oldhat{\mathbf{#1}}}

\title{Neural Machine Translation with 4-Bit Precision and Beyond}

\author{Alham Fikri Aji \& Kenneth Heafield \\
  School of Informatics, University of Edinburgh \\
  10 Crichton Street \\
  Edinburgh EH8 9AB \\
  Scotland, European Union \\
  \texttt{a.fikri@ed.ac.uk, kheafiel@inf.ed.ac.uk}
}

\date{July 2019}

\begin{document}

\maketitle
\begin{abstract}
    Neural Machine Translation (NMT) is resource intensive. We design a quantization procedure to compress NMT models better for devices with limited hardware capability. Because most neural network parameters are near zero, we employ logarithmic quantization in lieu of fixed-point quantization.  However, we find bias terms are less amenable to log quantization but note they comprise a tiny fraction of the model, so we leave them uncompressed.   
    We also propose to use an error-feedback mechanism during retraining, to preserve the compressed model as a stale gradient. We empirically show that NMT models based on Transformer or RNN architecture can be compressed up to 4-bit precision without any noticeable quality degradation. Models can be compressed up to binary precision, albeit with lower quality. The RNN architecture seems to be more robust to quantization, compared to the Transformer.
\end{abstract}
\section{Introduction}

Neural networks are becoming the state-of-the art for machine translation problems~\cite{wmt2018finding}. While neural machine translation (NMT)~\cite{bahdanau2014neural} yields better performance compared to its statistical counterpart, it is also more resource demanding. NMT embeds the tokens as vectors. Therefore the embedding layer has to store the vector representation of all tokens in both source and target vocabulary. Moreover, current state-of-the art architecture, such as Transformer or deep RNN usually requires multiple layers~\cite{vaswani2017attention, barone2017deep}.

Model quantization has been widely studied as a way to compress the model size and speed-up the inference process. However, most of these work were focused on convolution neural network for computer vision task~\cite{miyashita2016convolutional, lin2016fixed, hubara2016binarized, hubara2017quantized, jacob2018quantization}. A research in model quantization for NMT tasks is limited.

We first explore the use of logarithmic-based quantization over the fixed point quantization~\cite{miyashita2016convolutional}, based on the empirical findings that parameter distribution is not uniform~\cite{lin2016fixed, see2016compression}. Parameter's magnitude also varies across layers, therefore we propose a better way to scale the quantization centers. We also notice that biases do not consume noticeable amount of memory, therefore exploring the need to compress biases. Lastly, we explore the significance of retraining in model compression scenario. We adopt an error-feedback mechanism~\cite{1bitquant} to preserve the compressed model as a stale gradient, rather than discarding it every update during retraining.

\section{Related Work}

A considerable amount of research on model quantization has been done in the area of computer vision with convolutional neural networks; research on model quantization in the area of neural machine translation is much more limited. In this section, we will therefore also refer to work on neural models for image processing where appropriate.

\citet{lin2016fixed}, \citet{hubara2016binarized}, \citet{hubara2017quantized}, \citet{jacob2018quantization}, and \citet{junczys2018marian} all use linear quantization.
\citet{lin2016fixed} and \citet{hubara2016binarized} use a fixed scale parameter prior to model training; \citet{junczys2018marian} and \citet{jacob2018quantization} base it on the maximum tensor values for each matrix observed in the trained models.

Observing that their parameters are highly concentrated near 0 (see also \citealp{lin2016fixed,see2016compression}), \citet{miyashita2016convolutional} opt for logarithmic quantization. They report an improvement in preserving model accuracy over linear quantization while achieving the same model compression rate. 

\citet{hubara2017quantized} compress an LSTM-based architecture for language modeling to 4-bit without any quality degradation (while increase the unit size by a factor of 3). \citet{see2016compression} pruned an NMT model by removing any weight values that are lower than a certain threshold. They achieve 80\% model sparsity without any quality degradation.

The most relevant work with respect to our purposes is the submission of \citet{junczys2018marian} to the Shared Task on Efficient Neural Machine Translation in 2018. This submission applied an 8-bit linear quantization for NMT models without any noticeable deterioration in translation quality. 
Similarly, \citet{quinn2018pieces} proposed to use an 8-bit matrix multiplication to speedup an NMT system.

\section{Low Precision Neural Machine Translation}

\subsection{Log-Based Compression}
\label{algo}
\citet{lin2016fixed} and \citet{see2016compression} report that parameters in deep learning models are normally distributed: most of them are small values. To improve resolution for small values, where parameter density is the highest, we adopt logarithmic quantization following~\citet{miyashita2016convolutional}. 

We use the same quantization centers for positive and negative values.  When compressing to $B$ bits a single bit represents the sign while the remaining $B-1$ bits represent the log magnitude. The centers are tuned on the absolute value of the data.  

For efficient implementation and because the impact on quality was minimal after retraining, we use log base $2$. Log base $2$ means that exponentiation amounts to a bitshift while taking a log is equivalent to finding the leftmost $1$ in binary. We  find  that  tensors  might  not  have  the  same  magnitude. Therefore we also scale the quantization centers to approximate each tensor better. This approach is different to that of~\citet{miyashita2016convolutional}, where  quantization  centers  are  not-scaled,  thus  letting  every tensors to have the same centers. Formally, each quantization center takes the form $\pm S2^q$ where $S$ is a scaling factor and $q$ is an integer in the range $(-2^{B-1}, 0]$.  The scaling factor $S$ is selected separately for each tensor in the model.

To minimize the mean squared encoding error, values should be quantized to the nearest center. \citet{miyashita2016convolutional} find the nearest center in logarithmic space by taking the log then rounding to the nearest integer, which is not the same as finding the nearest center in normal space.  For example, this approach will quantize $5.8$ to $2^3$ instead of $2^2$, because $\log_2(5.8) \approx 2.536$ rounds to $3$. In normal space, $5.8$ rounds to $2^2$ not $2^3$.  

We can implement rounding to the nearest center in normal space efficiently by multiplying by $\frac{2}{3}$, taking the log, and rounding up to the next integer.  Let $x \in [2^q, 2^{q+1}]$.  In normal space,
\begin{equation}
\begin{split}
\text{$x$ rounds up to $2^{q+1}$}
\iff &x > \frac{2^q + 2^{q+1}}{2}\\
\iff&x > \frac{2^q(1 + 2)}{2} \\
\iff& \frac{2}{3}x > 2^q\\
\iff&\log_2{\frac{2}{3}x} > q\\
\end{split}
\end{equation}



Formally, the variable $v$ encodes as $(sign, q)$
\begin{equation}
\begin{split}
sign &= sign(v) \\
t & = clip(|v| / S, [1,2^{1-2^{B-1}}]) \\
q & = \lceil\log_2(\frac{2}{3}t)\rceil
\label{logquant}
\end{split}
\end{equation}
where $t$ is a temporary variable. This decodes to $v' \approx v$ as $v' = \text{sign} S 2^{q}$.  In practice, the sign is stored with $q$.  


\subsection{Selecting the Scaling Factor}
There are a few heuristics to choose a scaling factor $S$.  \citet{junczys2018marian} and \citet{jacob2018quantization} scale the model based on its maximum value, which can be very unstable especially during retraining. Alternatively, \citet{lin2016fixed} and \citet{hubara2016binarized} use a pre-defined step-size for fixed point quantization. Our objective is to select a scaling factor $S$ such that the quantized parameter is as close to the original as possible. Therefore, we optimize $S$ such that it minimizes the squared error between the original and the compressed parameter.

We propose a method to fit $S$ with Expectation-Maximization. We first start with an initial scale $S$ based on parameters' maximum value. For given $S$, we apply our quantization routine described in Equation~\ref{logquant} to a tensor $\vec{v}$, resulting in approximation $\vec{v'}$. For a given assignment $\vec{v'}$, we fit a new scale $S$ such that:

\begin{equation}
\label{eq-squareerror}
S = \arg\min_{S} \sum_{i}{(v'_i - v_i)^2}
\end{equation}

Substituting $v'_i$ within Eq.~\ref{eq-squareerror}, we have:

\begin{equation}
S = \arg\min_{S} \sum_{i}(sign(v_i) S 2^{q_i} - v_i)^2
\end{equation}

To simplify the equation, we can substitute $sign(v_i) 2^{q_i}$ as $a_i$, hence we have:

\begin{equation}
\label{eq-squareerrorsub}
S = \arg\min_{S} \sum_{i}(a_iS - v_i)^2
\end{equation}

To optimize the given objective, we take the first derivative of Equation~\ref{eq-squareerrorsub}  such that:

\begin{equation}
\begin{split}    
\frac{d}{dS} \sum_i(a_iS - v_i)^2 & = 0 \\
2 \sum_i (a_i (a_i S - v_i)) & = 0 \\
\sum_i (a_i^2 S) - \sum_i (a_i v_i) & = 0 \\
S \sum_i a_i^2 & = \sum_i (a_i v_i) \\
S & = \frac{\sum_i (a_i v_i)}{\sum_i {a_i}^2} \\
S & = \frac{\sum_i (sign(v_i) 2^{q_i} v_i)}{\sum_i (sign(v_i) 2^{q_i})^2 } \\
S & = \frac{\sum_i (2^{q_i} |v_i| )}{\sum_i 4^{q_i}}
\end{split}
\end{equation}

We optimize $S$ for each tensor independently.

\subsection{Retraining}

Unlike \citet{junczys2018marian}, we retrain the model after initial quantization to allow it to recover some of the quality loss. In the retraining phase, we compute the gradients normally with full precision. We then re-quantize the model after every update to the parameters, including fitting scaling factors. The re-quantization error is preserved in a residual valriable and added to the next step's parameter~\cite{1bitquant}. This error feedback mechanism was introduced in gradient compression techniques to reduce the impact of compression errors by preserving compression errors as stale gradient updates for the next batch~\cite{aji2017sparse, lin2017deep}. We see reapplying quantization after parameter updates as a form of gradient compression, hence we explore the usage of an error feedback mechanism to potentially improve the final model's quality.

\subsection{Handling Biases}

We do not quantize bias values in the model. We found that biases are not as highly concentrated near zero compared to other parameters. Empirically in our Transformer architecture, bias has higher standard deviation of 0.17, compared to 0.07 for other parameters. Attempting to log-quantize them used only a fraction of the available quantization points. In any case, bias values do not take up a lot of memory relative to other parameters. In our Transformer architecture, they account for only 0.2\% of the parameter values.


\subsection{Low Precision Dot Products}

Our end goal is to run a log-quantized model without decompressing it.  Activations coming into a matrix multiplication are quantized on the fly; intermediate activations are not quantized.  

We use the same log-based quantization procedure described in Section~\ref{algo}. However, we only attempt a max-based scale. Running the slower EM approach to optimize the scale before every dot product would not be fast enough for inference applications.  

The derivatives of ceiling and sign functions are zero almost everywhere and undefined in some places.  For retraining purposes, we apply a straight through estimator \cite{staightthrough} to the ceiling function.  For the sign function, we treat the quantization function differently for each individual value in $\vec{v}$, based on their sign. Therefore, we now compute $v'$ as:

\begin{equation}
v'=
    \begin{cases}
      2^q S , & \text{if}\ v > 0 \\
      -(2^q) S , & \text{otherwise}
    \end{cases}
\end{equation}

Since we multiply by a constant, the derivative will be either multiplied by $S$ or $-S$. We also have to compute $\frac{|v|}{S}$, which yeilds a derivative of $\text{sign}\frac{1}{S}$ which returns the chain derivative back to $1$. Hence, the derivative of our quantization function is:

\begin{equation}
  \frac{dv}{dv'}=
    \begin{cases}
      1, & \text{if}~1 \ge \frac{|v|}{S} \ge 2^{1-2^{B-1}} \\
      0, & \text{otherwise}
    \end{cases}  
\end{equation}

\section{Experiments}

\subsection{Experiment Setup}

We use systems for the WMT 2017 English to German news translation task for our experiment; these differ from the WNGT shared task setting previously reported. We use back-translated monolingual corpora~\cite{sennrich2015improving} and byte-pair encoding~\cite{sennrich2015neural} to preprocess the corpus. Quality is measured with BLEU~\cite{papineni2002bleu} score using sacreBLEU script~\cite{post2018call}.

We first pre-train baseline models with both Transformer and RNN architecture. Our Transformer model consists of six encoder and six decoder layers with tied embedding. Our deep RNN model consists of eight layers of bidirectional LSTM. Models were trained synchronously with a dynamic batch size of 40 GB per-batch using the Marian toolkit~\cite{junczys2018marian}. The models are trained for 8 epochs. Models are optimized with Adam~\cite{kingma2014adam}. The rest of the hyperparameter settings on both models are following the suggested configurations~\cite{vaswani2017attention,sennrich2017university}.

\subsection{4-bit Transformer Model}

In this first experiment we explore different ways to scale the quantization centers, the significance of quantizing biases, and the significance of retraining. We use pretrained Transformer model as our baseline, and apply our quantization algorithm on top of that.

\begin{table}[b!]
\centering
\begin{tabular}{lcccccc}
\hline
\bf Method & \multicolumn{1}{c}{\bf Scaling} & & & &   \\
 & \bf Fixed & \bf Max & \bf Optimized  & \bf Fixed & \bf Max & \bf Optimized \\
 \hline
 Baseline &  35.66 &  &  \\
 & \multicolumn{3}{c}{\textbf{Without retraining}} & \multicolumn{3}{c}{\textbf{With retraining}} \\
+ Model Quantization & 25.2 & 28.08 &  33.33 & 34.92 & 34.81 & 35.26 \\ 
+ No Bias Quantization & 34.16 & 34.29 & 34.31 & 35.09 & 35.25 & 35.47 \\ \hline
\end{tabular}
\caption{4-bit Transformer quantization performance for English to German translation, measured in BLEU score. We explore different method to find the scaling factor, as well as skipping bias quantization and retraining. }
\label{4-bit-abla}
\end{table}

Table~\ref{4-bit-abla} summarizes the results. Using a simple, albeit unstable max-based scaling has shown to perform better compared to the fixed quantization scale. However, fitting the scaling factor to minimize the quantization squared-error produces the best quality. Interestingly, the BLEU score differences between quantization centers are diminished after retraining.

We can also see improvements by not quantizing biases, especially without retraining. Without any retraining involved, we reached the highest BLEU score of 35.47 by using optimized scale, on top of uncompressed biases. Without biases quantization, we obtained 7.9x compression ratio (instead of 8x) with a 4-bit quantization. Based on this trade-off, we argue that it is more beneficial to keep the biases in full-precision.

Retraining has shown to improve the quality in general. After retraining, the quality differences between various scaling and biases quantization configurations are minimal. These results suggest that retraining helps the model to fine-tune under a new quantized parameter space.

To show the improvement of our method, we compare several compression approaches to our 4-bit quantization method with retraining and without bias quantization. One of arguably naive way to reduce model size is use smaller unit size. For Transformer, we set the Transformer feed-forward dimension to 512 (from 2048), and the embedding size to 128 (from 512). For RNN, we set the RNN dimension to 320 (from 1024) and the embedding size to 160 (from 512). This way, the model size for both architecture is relatively equal to the 4-bit compressed models.

We also introduce the fixed-point quantization approach as comparison, based on~\citet{junczys2018marian}. We made few modifications; Firstly We apply retraining, which is absent in their implementation. We also skip biases quantization. Finally, we optimize the scaling factor, instead of the suggested max-based scale.

\begin{table}[]
\centering
\begin{tabular}{lll}
\hline
\bf Method & \multicolumn{1}{c}{\bf Transformer} & \multicolumn{1}{c}{\bf RNN} \\
\hline
Baseline & 35.66 & 34.28 \\
Reduced Dimension & 29.03 \small (-6.63) & 30.88 \small (-3.40) \\
Fixed-Point Quantization & 34.61 \small  (-1.05) &  34.05 \small  (-0.23) \\
Ours & 35.47 \small (-0.19) & 34.22 \small (-0.06)\\
 \hline
\end{tabular}
\caption{Model performance (in BLEU) of various quantization approaches, on both Transformer and RNN architecture.}
\label{4-bit-compare}
\end{table}

Table~\ref{4-bit-compare} summarizes the result. It shown that reducing the model size by simply reducing the dimension performed worst. Logarithmic based quantization has shown to perform better compared to fixed-point quantization on both architecture. 

RNN model seems to be more robust towards the compression. RNN models have lesser quality degradation in all compression scenarios. Our hypothesis is that the gradients computed with a highly compressed model is very noisy, thus resulting in noisy parameter updates. Our finding is in line with prior research which state that Transformer is more sensitive towards noisy training condition~\cite{chen2018best, aji2019making}.

\subsection{Quantized Dot-Product}

We now apply logarithmic quantization for all dot-product inputs. We use the same quantization procedure as the parameter, however we do not fit the scaling factor as it is very inefficient. Hence, we try using max-scale and fixed-scale. For the parameter quantization, We use optimized scale with uncompressed biases, based on the previous experiment.

\begin{table}[]
\centering
\begin{tabular}{lll}
\hline
\bf Method & \bf Transformer & \bf RNN \\
\hline
Baseline & 35.66 & 34.28 \\
 + Model Quantization & 35.47 \small (-0.19) & 34.22 \small  (-0.06) \\
 + Dot Product Quantization & 35.05 \small (-0.61) & 33.12 \small (-1.16)\\
 \hline
\end{tabular}
\caption{Model performance (in BLEU) of model quantization with dot-product quantization, on both Transformer and RNN architecture.}
\label{quantact}
\end{table}

Table~\ref{quantact} shows the quality result of the experiment. Generally we see a quality degradation compared to a full-precision dot-product. There is no significant difference between using max-scale or fixed-scale. Therefore, using fixed-scale might be beneficial, as we avoid extra computation cost to determine the scale for every dot-product operations.



\subsection{Beyond 4-bit precision}

With 4-bit quantization and uncompressed biases, we obtain 7.9x compression rate. Bit-width can be set below 4 bit to obtain an even better compression rate, albeit introducing more compression error. To explore this, we sweep several bit-width. We skip bias quantization and optimize the scaling factor.

\begin{table}[]
\centering
\begin{tabular}{rllll}
\hline
\bf Bit & \multicolumn{2}{c}{\bf Transformer} & \multicolumn{2}{c}{\bf RNN} \\ 
& \multicolumn{1}{c}{\bf Size (rate)} & \multicolumn{1}{c}{\bf BLEU($\Delta$)} & \multicolumn{1}{c}{\bf Size (rate)} & \multicolumn{1}{c}{\bf BLEU($\Delta$)}  \\
\hline
32 & 251 MB   & 35.66 & 361 MB & 34.28 \\
4 &  \phantom{0}32 MB  \small (\phantom{0}7.88x) & 35.47 \small (-0.19) & \phantom{0}46 MB  \small (\phantom{0}7.90x) & 34.22 \small (-0.06) \\

3 &  \phantom{0}24 MB  \small (10.45x) & 34.95 \small (-0.71) & \phantom{0}34 MB  \small (10.49x) & 34.11 \small (-0.17) \\

2 &  \phantom{0}16 MB  \small (15.50x) & 33.40 \small (-2.26) & \phantom{0}23 MB  \small (15.59x) & 32.78 \small (-1.50) \\

1 &  \phantom{00}8 MB \small (30.00x) & 29.43 \small (-6.23) & \phantom{0}12 MB  \small (30.35x) & 31.71 \small (-2.51) \\
 \hline
\end{tabular}
\caption{Compression rate and performance of both Transformer and RNN with various bit-widths. The compression rate between Transformer and RNN is not equal, as they have different bias to parameter size ratio.}
\label{compression-rate}
\end{table}

Training an NMT system below 4-bit precision is still a challenge. As shown in Table~\ref{compression-rate}, model performance degrades with less bit used. While this result might still be acceptable, we argue that the result can be improved. One idea that might be interesting to try is to increase the unit-size in extreme low-precision setting. We shown that 4-bit precision performs better compared to full-precision model with (near) 8x compression rate. In addition, \citet{han2015deep} has shown that 2-bit precision image classification can be achieved by scaling the parameter size. Alternative approach is to have different bit-width for each layers~\cite{hwang2014fixed, anwar2015fixed}.

We can also see RNN robustness over Transformer in this experiment, as RNN models degrade less compared to the Transformer counterpart. RNN model outperforms Transformer when compressing at binary precision.

\section{Conclusion}

We compress the model size in neural machine translation to approximately 7.9x smaller than 32-bit floats by using a 4-bit logarithmic quantization. Bias terms behave different and can be left uncompressed without affecting the compression rate significantly. We also find that retraining after quantization is necessary to restore the model's performance.

\bibliographystyle{acl_natbib.bst}
\bibliography{ref}
\end{document}